\documentclass[conference]{IEEEtran}
\IEEEoverridecommandlockouts
\usepackage{cite}
\usepackage{amsmath,amssymb,amsfonts,commath}
\usepackage{algorithm}
\usepackage{algorithmic}
\usepackage{graphicx}
\usepackage{textcomp}
\usepackage{xcolor}
\usepackage{amsthm}
\usepackage{regexpatch,fancyvrb,xparse}

\usepackage{mathtools}

\DeclarePairedDelimiter\floor{\lfloor}{\rfloor}

\def\BibTeX{{\rm B\kern-.05em{\sc i\kern-.025em b}\kern-.08em
    T\kern-.1667em\lower.7ex\hbox{E}\kern-.125emX}}
\begin{document}

\title{Network Pruning via Annealing and Direct Sparsity Control}

\author{\IEEEauthorblockN{Yangzi Guo}
\IEEEauthorblockA{\textit{Department of Mathematics} \\
\textit{Florida State University}\\
Tallahassee, Florida, USA \\
yguo@math.fsu.edu}
\and
\IEEEauthorblockN{Yiyuan She}
\IEEEauthorblockA{\textit{Department of Statistics} \\
\textit{Florida State University}\\
Tallahassee, Florida, USA \\
yshe@stat.fsu.edu}
\and
\IEEEauthorblockN{Adrian Barbu}
\IEEEauthorblockA{\textit{Department of Statistics} \\
\textit{Florida State University}\\
Tallahassee, Florida, USA \\
abarbu@stat.fsu.edu}
}

\maketitle

\begin{abstract}
Artificial neural networks (ANNs) especially deep convolutional neural networks are very popular these days and have been proved to successfully offer quite reliable solutions to many vision problems. 
However, the use of deep neural networks is widely impeded by their intensive computational and memory cost. 
In this paper, we propose a novel efficient network pruning framework that is suitable for both non-structured and structured channel-level pruning. Our proposed method tightens a sparsity constraint by gradually removing network parameters or filter channels based on a criterion and a schedule. 
The attractive fact that the network size keeps dropping throughout the iterations makes it suitable for the pruning of any untrained or pre-trained network. 
Because our method uses a $L_0$ constraint instead of the $L_1$ penalty, it does not introduce any bias in the training parameters or filter channels. 
Furthermore, the $L_0$ constraint makes it easy to directly specify the desired sparsity level during the network pruning process. 
Finally, experimental validation on extensive synthetic and real vision datasets show that the proposed method obtains better or competitive performance compared to other states of art network pruning methods.
\end{abstract}

\vspace{+12mm}
\section{Introduction}
In recent years, artificial neural networks (ANNs) especially deep convolutional neural networks (DCNNs) are widely applied and have become the dominant approach in many computer vision tasks. 
These tasks include image classification \cite{krizhevsky2012imagenet,simonyan2014very,he2016deep,huang2017densely}, object detection \cite{girshick2014rich,ren2015faster}, semantic segmentation \cite{long2015fully}, 3D reconstruction \cite{dou2017end}, etc. 
The quick development in the deep learning field leads to network architectures that can go nowadays as deep as 100 layers and contain millions or even billions of parameters.
Along with that, more and more computation resources must be utilized to successfully train such a deep modern neural network.

The deployment of DCNNs in real applications is largely impeded by their intensive computational and memory cost. 
With this observation, the study of network pruning methods that learn a smaller sub-network from a large original network without losing much accuracy has attracted a lot of attention. 
Network pruning algorithms can be divided into two groups: non-structured pruning and structured pruning. 
The earliest work for non-structured pruning is conducted by \cite{lecun1990optimal}, the most recent work is done by \cite{han2015deep,han2015learning}. 
The non-structured pruning aims at directly pruning parameters regardless of the consistent structure for each network layer. 
This renders modern GPU acceleration technique unable to obtain computational benefits from the irregular sparse distribution of parameters in the network, only specialized software or hardware accelerators can gain memory and time savings. 
The advantage of non-structured pruning is that it can obtain high network sparsity and at the same time preserve the network performance as much as possible. 
On the other side, structured pruning aims at directly removing entire convolutional filers or filter channels. 
Li \textit{et al.} \cite{li2016pruning} determines the importance of a convolutional filter by measuring the sum of its absolute weights. 
Liu \textit{et al.} \cite{liu2017learning} introduces a $L_1$-norm constraint in the batch normalization layer to remove filter channels associated with smaller $\gamma$. 
Although structured pruning cannot obtain the same level of sparsity as non-structured pruning, it is more friendly to modern GPU acceleration techniques and independent of any specialized software or hardware accelerators.

Unfortunately, many of the existing non-structured and structured pruning techniques are conducted in a layer-wise way, requiring a sophisticated procedure for determining the hyperparameters of each layer in order to obtain a desired number of weights or filters/channels in the end. 
This kind of pruning manner is not effective nor efficient. 

We combine regularization techniques with sequential algorithm design and direct sparsity level control to bring forward a novel network pruning scheme that could be suitable for either non-structured pruning or structured pruning  (particular for filter channel-wise pruning of DCNNs with Batch Normalization layers). 
We investigate a parameter estimation optimization problem with a $L_0$-norm constraint in the parameter space, together with the use of annealing to lessen the greediness of the pruning process and a general metric to rank the importance of the weights or filter channels. 
An attractive property is that parameters or filter channels are removed while the model is updated at each iteration, which makes the problem size decrease during the iteration process. 
Experiments on extensive real vision data, including the MNIST, CIFAR, and SVHN provide empirical evidence that the proposed network pruning scheme obtains a performance comparable to or better than other state of art pruning methods.

\section{Related Work}

Network pruning is a very active research area nowadays, it provides a powerful tool to accelerate the network inference by having a much smaller sub-network without too much loss in accuracy. 
The earliest work about network pruning can be dated back to 1990s, when \cite{lecun1990optimal} and  \cite{hassibi1993second} proposed a weight pruning method that uses the Hessian matrix of the loss function to determine the unimportant weights. 
Recently, \cite{han2015learning} used a quality parameter multiplied by the standard deviation of a layer's weights to determine the pruning threshold. 
A weight in a layer will be pruned if its absolute value is below that threshold. 
\cite{guo2016dynamic} proposed a pruning method that can properly incorporate connection slicing into the pruning process to avoid incorrect pruning. 
These pruning schemes mentioned above are all non-structured pruning, needing specialized hardware or software to gain computation and time savings.

For structured pruning, there are also quite a few works in the literature. 
\cite{li2016pruning} determine the importance of a convolutional filter by measuring the sum of its absolute weights. 
\cite{hu2016network} compute the average percentage of zero activations after the ReLu function and determine to prune the corresponding filter if its this percentage score is high. 
\cite{he2017channel} propose an iterative two-step channel pruning method by a LASSO regression based channel selection and least square reconstruction. 
\cite{liu2017learning} introduce a $L_1$-norm constraint in the batch normalization layer to remove filter channels associated with smaller $|\gamma|$. 
\cite{zhou2018online} impose an extra cluster loss term in the loss function that forces filters in each cluster to be similar and only keep one filter in each cluster after training. 
\cite{yu2018nisp} utilize a greedy algorithm to perform channel selection in a layer-wise way by constructing a specific optimization problem.

\section{Network Pruning via Annealing and Direct Sparsity Control}

Given a set of training examples $\mathcal{D} = \{ (\mathbf{x_{i}},y_i), i=1,...,N \}$ where $\mathbf{x}$ is an input and y is a corresponding target output, with a differentiable loss function $L(\cdot)$ we can formulate the pruning problem for a neural network with parameters $\mathcal{W} = \{ \mathbf{(W_{j}},\mathbf{b_j}), j=1,...,L \}$ as following \textbf{constrained} problem
\begin{alignat}{1}
\min_{\mathcal{W}} \quad & L(\mathcal{W}) \ \ \ \ \ \ \ \ \ \ \ \mbox{s.t.}\quad ||\mathcal{W}||_0 \leq K
\label{cons_weight}
\end{alignat}
where the $L_0$ norm bounds the number of non-zero parameters in $\mathcal{W}$ to be less than or equal to a specific positive integer $K$. 

For non-structured pruning, we directly address the pruning problem in the whole $\mathcal{W}$ space. The final $\mathcal{W}$ will have an irregular distribution pattern of the zero-value parameters across all layers. 

For structured pruning, suppose the DCNN is with convolutional filters or channels $\mathcal{C} = \{C_j,j=1,...,M \}$, we can replace the constrained problem (\ref{cons_weight}) by 
\begin{alignat}{2}
\min_{\mathcal{W}} \quad & L(\mathcal{W}) \ \ \ \ \ \ \ \ \ \ \ \mbox{s.t.}\quad ||\mathcal{C}||_0 \leq K
\label{cons_channel}
\end{alignat}
By solving the problem (\ref{cons_channel}), we will obtain the $\mathcal{W}$ on the convolutional layers having more uniform zero-value parameter distribution, specialized in some filters or filter channels. 

These constrained optimization problems (\ref{cons_weight}) and (\ref{cons_channel}) facilitate parameter tuning because our sparsity parameter $K$ is much more intuitive and easier to specify in comparison to penalty parameters such as $\lambda$ in $\lambda||\mathcal{W}||_1$ and $\lambda||\mathcal{C}||_1$. 

In this work, we will focus on the study of the weight-level pruning (non-structured pruning) for all neural networks and channel-level pruning (structured pruning) particularly for neural networks with Batch Normalization layers.

\section{Basic Algorithm Description}

Some key ideas in our algorithm design are: 
a) We conduct our pruning procedures in the specified parameter spaces; 
b) We use an annealing plan to directly control the sparsity level in each parameter space; 
c) We gradually remove the most "unimportant" parameters or channels to facilitate computation. 
The prototype algorithms, summarized in Algorithm \ref{alg:DSC_1} and \ref{alg:DSC_2}, show our ideas.
It starts with either an untrained or pre-trained model and alternates two basic steps: one step of parameter updates towards minimizing the loss $L(\cdot)$ by gradient descent and one step that removes some parameters or channels according to a ranking metric $\mathcal{R}$.

\begin{algorithm}[htb]
	\caption{{\bf Network Pruning via Direct Sparsity Control - Weight-Level (DSC-1)}}
	\label{alg:DSC_1}
	\begin{algorithmic}
		\STATE {\bfseries Input:} Training set $T=\{(\mathbf{x_i},y_i)\}_{i=1}^{n}$, desired parameter space $\{\mathcal{W}_{j} | \cup\mathcal{W}_{j}=\mathcal{W} \ \& \cap\mathcal{W}_{j}=\varnothing\}_{j=1}^{B}$,
		desired number $\{K_j\}_{j=1}^{B}$ of parameters, desired annealing schedule $\{M_{j}^{e},e=1,..,N^{iter}\}_{j=1}^{B}$, an ANN model.
		\STATE {\bfseries Output:} Pruned ANN depending on exactly $\{K_j\}_{j=1}^{B}$ parameters in each parameter space $\{\mathcal{W}_{j}\}_{j=1}^{B}$.
	\end{algorithmic}
	\begin{algorithmic} [1]
		\STATE If the ANN is not pre-trained, train it to a satisfying level.
		\FOR {$e = 1$ to $N^{iter}$}
			\STATE Sequentially update $\mathcal{W} \leftarrow \mathcal{W} - \eta\frac{\partial L(\mathcal{W})}{\partial \mathcal{W}}$ via backpropagation.
			\FOR {$j=1$ to $B$}
			    \STATE Keep the $M_{j}^{e}$ most important parameters in $\mathcal{W}_{j}$ based on ranking metric $\mathcal{R}$.
			\ENDFOR
		\ENDFOR
		\STATE Fine-tune the pruned ANN with exactly $\{K_j\}_{j=1}^{B}$ parameters in each parameter space $\{\mathcal{W}_{j}\}_{j=1}^{B}$.
\end{algorithmic}
\end{algorithm}

The intuition behind our \textbf{DSC} algorithms is that during the pruning process, each time we remove a certain number of the most unimportant parameters/channels in each parameter/channel space based on an annealing schedule. This ensures that we do not inject too much noise in the parameter/channel dropping step so that the pruning procedure can be conducted smoothly. Our method directly controls the sparsity level obtained at each parameter/channel space, unlike many layer-wise pruning methods where a sophisticated procedure has to be used to control how many parameters are kept, because pruning the weights or channels in all layers simultaneously can be very time-consuming. 

\begin{algorithm}[htb]
	\caption{{\bf Network Pruning via Direct Sparsity Control - Channel-Level (DSC-2)}}
	\label{alg:DSC_2}
	\begin{algorithmic}
		\STATE {\bfseries Input:} Training set $T=\{(\mathbf{x_i},y_i)\}_{i=1}^{n}$, desired channel space $\{\mathcal{C}_{j} | \cup\mathcal{C}_{j}=\mathcal{C} \ \& \cap\mathcal{C}_{j}=\varnothing\}_{j=1}^{B}$,
		desired number $\{K_j\}_{j=1}^{B}$ of channels, desired annealing schedule $\{M_{j}^{e},e=1,..,N^{iter}\}_{j=1}^{B}$, a DCNN model.\STATE {\bfseries Output:} Pruned DCNN depending on exactly $\{K_j\}_{j=1}^{B}$ channels in each channel space $\{\mathcal{C}_{j}\}_{j=1}^{B}$.
	\end{algorithmic}
	\begin{algorithmic} [1]
		\STATE If the DCNN is not pre-trained, train it to a satisfying level.
		\FOR {$e = 1$ to $N^{iter}$}
			\STATE Sequentially update $\mathcal{W} \leftarrow \mathcal{W} - \eta\frac{\partial L(\mathcal{W})}{\partial \mathcal{W}}$ via backpropagation
			\FOR {$j=1$ to $B$}
			    \STATE Keep the $M_{j}^{e}$ most important channels in $\mathcal{C}_{j}$ based on ranking metric $\mathcal{R}$.
			\ENDFOR
		\ENDFOR
		\STATE Fine-tune the pruned DCNN with exactly $\{K_j\}_{j=1}^{B}$ channels in each parameter space $\{\mathcal{C}_{j}\}_{j=1}^{B}$.
\end{algorithmic}
\end{algorithm}

Through the annealing schedule, the support set of the network parameters or channels is gradually shrunken until we reach $||\mathcal{W}||_0\leq K$ or  $||\mathcal{C}||_0\leq K$. 
The keep-or-kill rule is based on the ranking metric $\mathcal{R}$ and does not involve any information of the objective function $L$. This is in contrast to many ad-hoc networking pruning approaches that have to modify the loss function and can not easily be scaled up to many existing pre-trained models.

\section{Implementation Details}
In this part, we provide implementation details of our proposed \textbf{DSC} algorithms. 

\begin{figure}[t]
\centering
\includegraphics[width=7.5cm]{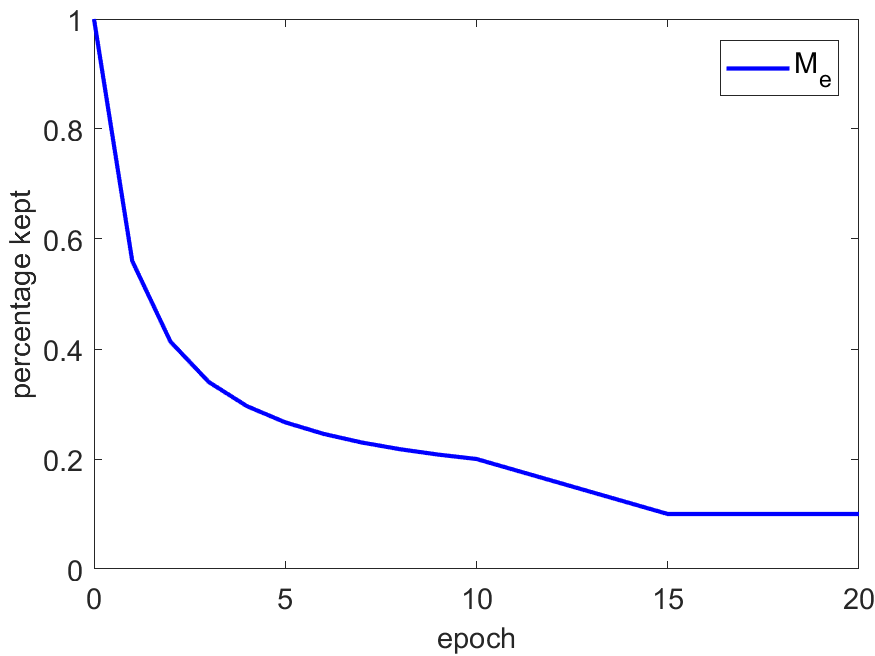}
\caption{Annealing schedule with $N_1=10,N^{iter}=20, N^c=1$, $p_0=0.8, p=0.9, \nu=0.02$.}\label{fig:schedule}
\end{figure}

First, the annealing schedule $M_e$ is determined empirically. Our experimental experience shows that the following annealing plans can perform well to balance the efficiency and accuracy:
$$
M_e=\left\{
\begin{aligned}
& (1-p_{0})+p_{0}(\frac{N_{1}-e}{\mu e + N_{1}}))M,   \ \ \ 1 \leq e < N_{1}      \\
& (1 - \min(p,p_{0}+\floor*{\frac{e-N_1}{N^{c}}}\nu))M,  N_1 < e \leq N^{iter} \\
\end{aligned}
\right.
\label{Me}
$$
Here $M$ is the total number of parameters or channels in the neural network. Our $M_e$ consists two parts. 
The first part can be used to quickly prune the unimportant parameters with a reasonable value of $\mu$ down to a percentage $p_0$ of the parameters. 
The second part can further refine our pruned sub-network to a more compact model. 
$\mu$ is the pruning rate and we will set it to $\mu=10$ for all experiments. 
$p_0 \in [0,1]$ denotes the percentage of parameters or channels to be pruned in the first part. 
$p \in [0,1]$ denotes the final pruning percentage goal at the end of the pruning procedure, thus the number of remaining parameters is $K=M(1-p)$. 
The parameter $N^{c}$ specifies how many epochs to train before performing another pruning. 
We will select $N^c \in \{1,2\}$. $\nu$ denotes the incremental pruning percentage as the annealing continues and will be set to $\nu\in \{0.005,0.01,0.02\}$. 
An example of an annealing schedule $M_e$ (with $M=1$ for clarity) with $N_1=10,N^{iter}=20, N^c=1$, $p_0=0.8, p=0.9$, and $ \nu=0.02$ is shown in Figure \ref{fig:schedule}.

Second, as the convolutional layers and fully connected layers have very different behavior in a DCNN, we will prune them separately during the structured and non-structured pruning process, i.e. we will fix the convolutional layer parameters while pruning the fully connected layer, and vice versa.

Third, the ranking metric $\mathcal{R}$ we select for structured and non-structured pruning is different. 
For non-structured pruning, the parameter dropping procedure based on the magnitude of the parameter yields quite good pruning results in our experiments. 
Therefore we will select it as our metric to rank the importance of parameters for all our non-structured pruning experiments:
\begin{equation}
\begin{aligned}
\mathcal{R}(w) = |w|, w \in \mathcal{W}
\label{R_structured}
\end{aligned}
\end{equation}
For structured channel pruning, various dropping criteria are proposed. 
One family of channel pruning metrics are based on the value of the channel weights. Li
\cite{li2019exploiting} uses the $L_1$-norm by summing up the magnitude of all channel weights to rank the importance of the metric in a channel space; Wen \cite{wen2016learning} suggests the use of the $L_2$-norm. 
Another family of channel pruning metrics \cite{liu2017learning} lies in the absolute value of the Batch Normalization scales, as Batch normalization
\cite{ioffe2015batch} has been widely adopted by most modern DCNNs to accelerate the training speed and convergence. Assume $z_{in}$ and $z_{out}$ to be the input and output of a Batch Normalization layer, we can formulate the transformation of that BN layer performs as:
$$
BN(z_{in}) = \frac{z_{in}-\mu_\mathbf{B}}{\sqrt{\sigma^{2}_{\mathbf{B}}+\epsilon}}; z_{out}=\gamma\cdot BN(z_{in}) + \beta
\label{BN_scale}
$$
where $\mathbf{B}$ denotes the mini-batch statistic of input activations, $\mu_\mathbf{B}$ and $\sigma_{\mathbf{B}}$ are the mean and standard deviation over $\mathbf{B}$, $\gamma$ and $\beta$ are trainable scale and shift parameters of the affine transformation. Liu \cite{liu2017learning} directly leverages the parameters $\gamma$ in the Batch Normalization layers as the scaling factors they need for channel pruning. They impose a $L_1$ norm on each Batch Normalization layer for the $\gamma$ to reformulate the training loss function. Here we combine the metrics of these two families to enjoy wider flexibility on the DCNNs and define our ranking metrics as follows:
\begin{equation}
\begin{aligned}
 \mathcal{R}_{\mathbf{B}}(\mathcal{C}) & = |\gamma_\mathcal{C}|  \\ 
 \mathcal{R}_{\mathbf{L}}(\mathcal{C}) & =  (||\mathcal{C}||_{L_1} + ||\mathcal{C}||_{L_2})/2 \\
 \mathcal{R}(\mathcal{C})              & = 
 \alpha\cdot\frac{\mathcal{R}_{\mathbf{B}}(\mathcal{C})}{\mathcal{R}_{\mathbf{B}}^{max}(\mathcal{C})} + (1-\alpha)\cdot\frac{\mathcal{R}_{\mathbf{L}}(\mathcal{C})}{\mathcal{R}_{\mathbf{L}}^{max}(\mathcal{C})}
\end{aligned}
\label{eq:R_structured}
\end{equation}
where $\gamma_\mathcal{C}$ is the scale parameter of the BN for channel $\mathcal{C}$, $\alpha\in[0,1]$ is a hyper-parameter that needs to be specified to balance the two ranking terms $\mathcal{R}_{\mathbf{B}}$ and $\mathcal{R}_{\mathbf{L}}$. 
The main differences from the other pruning methods are that we do not make any modifications to the loss function, but utilize a $L_0$ norm constraint and we use an annealing schedule to gradually eliminate channels and lessen the greediness.

Fourth, after the pruning process, we will conduct a fine-tuning procedure to gain back the performance lost during the pruning period. 
Before we start the fine-tuning, we can remove for non-structured pruning the neurons that have zero incoming or outgoing degree and structured-pruning the convolution filter channels with all zero parameters to form a more compact network for later inference use. 

\section{Experiments}

In this section, we first present a simulation on a synthetic dataset named parity dataset \cite{zhang2017learnability} to demonstrate the effectiveness of our \textbf{DSC} algorithm with selected $M_e$ annealing plan. Then we conduct non-structured pruning with Lenet-300-100 and LeNet-5 \cite{lecun1998gradient} on MNIST \cite{lecun-mnisthandwrittendigit-2010} dataset. Finally we conduct our experiments with VGG-16 \cite{simonyan2014very} and DenseNet-40 \cite{huang2017densely} on CIFAR \cite{krizhevsky2009learning} and SVHN \cite{netzer2011reading} dataset for structured channel pruning.

\subsection{Synthetic Parity Dataset}
The parity data with noise is a classical problem in computational learning theory \cite{zhang2017learnability}. 
The data has feature vector $\mathbf{x}\in {\mathbb R}^p$ which is uniformly drawn from $\{-1,+1\}^p$. 
The label is generated follow the XOR logic: for some unknown subset of $k$ indices $1\leq i_1 < ... < i_k$, the label value is set as
$$ y=\left\{
\begin{aligned}
& x_{i_1}x_{i_2}...x_{i_k} \ \ \ \ \ \ \ \ \ \text{with probablity} \ 0.9\\
& -x_{i_1}x_{i_2}...x_{i_k} \ \ \ \ \ \text{with probablity} \ 0.1\\
\end{aligned}
\right.
$$
That is this dataset cannot be perfectly separated and the best classifier would have a prediction error of 0.1.

\begin{figure}[t]
\centering
\includegraphics[width=7.5cm]{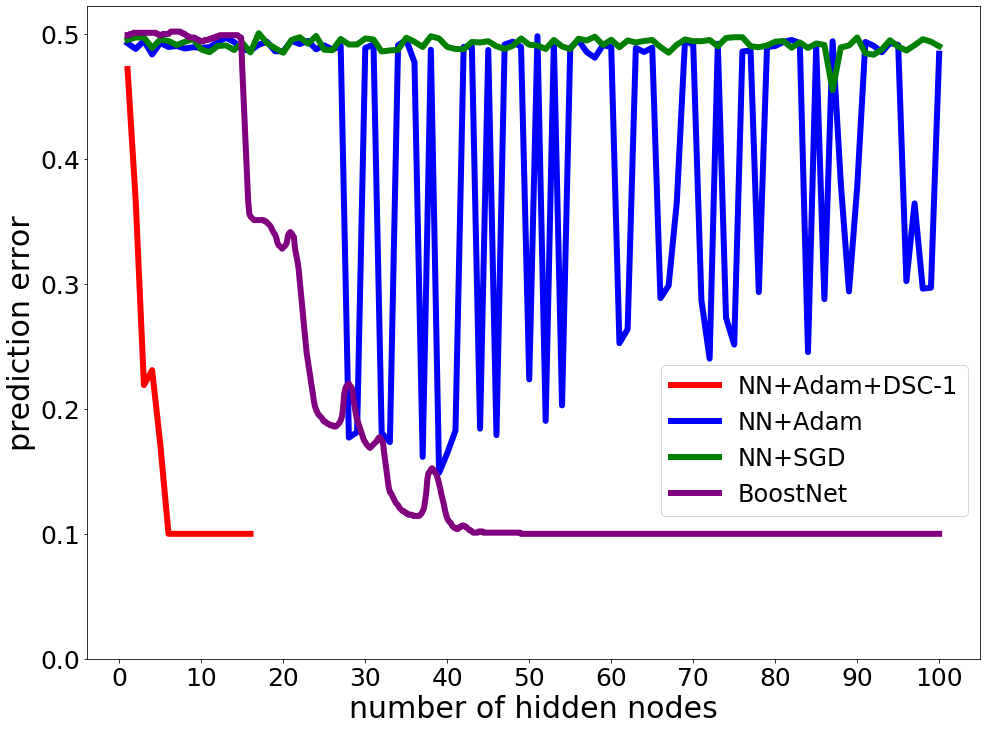}
\caption{Test error vs number of hidden nodes. Comparison between single hidden layer neural networks trained by NN + Adam + \textbf{DSC-1} starting with 256 hidden nodes, NN + Adam, NN + SGD and BoostNet.}\label{fig:parity}
\end{figure}

This kind of dataset is frequently used to test different optimizers and regularization techniques on the neural network (NN) model. We perform the experiment in $p=50$ dimensional data with parities $k=5$. 
The training set, valid set, and testing set contain respectively 15K, 5K and 5K data points. 
We train a one hidden layer neural network with default stochastic gradient descent (SGD) optimizer, Adam \cite{kingma2014adam} optimizer and Adam + \textbf{DSC-1}. 
For NN with Adam + \textbf{DSC-1}, we start with $256$ hidden nodes, and down to a hidden node number $B$ in the range $B\in [1,16]$ using annealing schedule $M_e$. We report the best result out of 10 independent random initializations. Recently, a neural network based boosting method named BoostNet \cite{zhang2017learnability} significantly outperformed a normal NN on this data. As Zhang's \cite{zhang2017learnability} experiment setting is very similar to us but with more training data, so we directly extract their results and report together with our experimental outcomes. The comparison of the test errors is shown in Figure \ref{fig:parity}. 

We can see that the NN with the SGD optimizer cannot learn any good model with less than 100 hidden nodes on this data, while a NN with the Adam optimizer can learn some pattern when the number of hidden nodes is greater than 25, but still mostly cases are trapped in shallow local optima. 
The BoostNet can learn well if the hidden node number is greater than about 45 hidden nodes. 
The best performance is achieved by NN with Adam + \textbf{DSC-1}, with 256 starting hidden nodes. After applying \textbf{DSC-1} during the NN training, we only needed to keep as few as 6 hidden nodes to get the best possible prediction error. This observation implies: The \textbf{DSC-1} algorithm has a good capability to find a global or deep enough local optimum by gradually removing unimportant connections; The direct sparsity control design can help the final NN model reach very close to the most compact model achievable.

\subsection{Non-structured Pruning on MNIST}

The MINIST dataset provided by \cite{lecun-mnisthandwrittendigit-2010} is a handwritten digits dataset that is widely used in evaluating machine learning algorithms. 
It contains 50K training observations, 10K validation and 10K testing observations respectively. 
In this section, we will test our non-structured pruning method \textbf{DSC-1} on two network models: LeNet-300-100 and LeNet-5.  

LeNet-300-100 \cite{lecun1998gradient} is a classical fully connected neural network with two hidden layers. The first hidden layer has 300 neurons and the second has 100. The LeNet-5 is a conventional convolutional neural network that has two convolution layers and two fully connected layers. 
LeNet-300-100 consists of 267K learnable parameters and LeNet-5 consists of 431K. To have a fair comparison with \cite{han2015learning}, we follow the same experimental setting by using the default SGD method, training batch size and initial learning rate to train the two models from scratch. 
After a model with similar performance was obtained, we stop the training and directly apply our \textbf{DSC-1} pruning algorithm to compress the model. 
During the pruning and retraining procedure, a learning rate with 1/10 of the original network's learning rate is adopted. 
A momentum with value of 0.9 is used to speed up the model retraining. 

\begin{table}[ht]
\centering
\begin{tabular}{llll}
\hline
\hline
Model  & Error & Params &Prune Rate   \\
\hline
Lenet-300-100 (Baseline)  &1.64\% &267k &- \\
Lenet-300-100 (Han \textit{et al.}) &1.59\% &22K &91.8\% \\
Lenet-300-100 (Ours) &\bf{1.57\%} &\bf{17.4K} &\bf{93.5\%} \\
\hline
Lenet-5 (Baseline) &0.8\% &431K &- \\
Lenet-5 (Han \textit{et al.}) &0.77\% &36k &91.6\% \\
Lenet-5 (Ours) &0.77\% &\bf{15.8k} &\bf{96.4\%} \\
\hline
\hline
\end{tabular}
\caption{Non-structured pruning comparison. Our \textbf{DSC-1} pruning method can learn a more compact sub-network.}
\label{tab:MNIST_1}
\end{table}

\begin{table}[htb]
\centering
\begin{tabular}{lllll}
\hline
\hline
Model  & Layer & Params. &Han \% &Ours \%   \\
\hline
                &fc1   &236K  &8\%   &4.6\%    \\
Lenet-300-100   &fc2   &30K   &9\%   &20.1\%   \\
                &fc3   &1K    &26\%  &68.5\%  \\
                &Total &267K  &8.2\% &6.5\%    \\
\hline
                &conv1 &0.5K &66\% &75\% \\
                &conv2 &25K &12\% &29.1\% \\
Lenet-5         &fc1   &400K &8\% &1.8\% \\
                &fc2   &5K &19\% &17.2\% \\
                &Total &431K &8.4\% &3.6\% \\
\hline
\hline
\end{tabular}
\caption{Layer by layer compression comparisons on LeNet-300-100 and LeNet-5. The percentage of remaining parameters of \cite{han2015learning}'s pruning method is displayed in the third column, our \textbf{DSC-1} pruning is displayed in the last column.}
\label{tab:MNIST_2}
\end{table}

In LeNet-300-100, a total of 20 epochs were used for both pruning and fine-tuning. 
For the annealing schedule, $p_0$ is directly set to 0.85 without using any annealing schedule. 
Then we follow the fine-grain pruning annealing schedule which $N^c=1$ and $\nu=0.05$ to reach at the final percentage goal $p=0.935$. 

The remaining epochs are used for fine-tuning purposes. 
In LeNet-5, the pruning for fully connected layers and convolutional layers are treated separately. 
For pruning on fully connected layers, we directly set at $p_0=0.9$ and then reach $p=0.98$ with $N^c=1,\nu=0.05$. 
For the convolutional layers we start with $p_0=0$, $N^{c}=1$ and $\nu = 0.05$ to reach at $p=0.7$. 
The total number of pruning and retraining epochs for LeNet-5 is 40 epochs.
After several experimental trials, we output our best result in Table \ref{tab:MNIST_1} .

From the result table shown above, one can observe that our proposed non-structured pruning algorithm can learn a more compact sub-network for both LeNet300-100 and LeNet-5 with comparable performance with  \cite{han2015learning}. 

By using a hyperparameter we can directly control the sparsity level to get close to the most compact model achievable. 
It is not hard to conjecture that using a quality factor times the variance as a pruning threshold in each layer as proposed by \cite{han2015learning} cannot exactly determine how many parameters should be kept. 
Our method can directly control the sparsity level and therefore enjoy a higher possibility to reach the position of the most compact sub-network. 

Table \ref{tab:MNIST_2} shows the layer-by-layer compression comparisons between ours and \cite{han2015learning}. It is interesting to see that although two different pruning algorithms yield a similar performance result, the network architecture is quite different. 
Our \textbf{DSC-1} algorithm controls the directly specified sparsity level in the parameter space with an annealing schedule, this ensures the target sub-network can learn its pattern in an automatic way. 
For LeNet300-100, the most parameter killing comes from the first layer, which is quite reasonable as the images in the MNIST dataset are grayscale containing a large portion of pure black pixels. 
This large portion of black pixels almost has nothing to contribute to the neural network learning of useful information. 
The least parameter percentage dropping comes from the output layer, preserved as high as $68.5\%$. 
We can conjecture the reason for this behavior could be that as the most unrelated features are removed from the first fully connected layer, the output layer should remain a considerable number of parameters to bear the weight of those kept and useful features. 
For LeNet-5, the most parameter preservation occurs in the first convolutional layer. This is again really very reasonable, as indeed the first layer should be the most important layer that directly extracts relevant features from the raw input image pattern. 
Our direct sparsity control strategy lets the network itself decide which part is more important, and which part contains most irrelevant or junk connections that could be removed safely. 
The parameter percentage distribution of the two fully connected layers in LeNet-5 has a similar behavior as in LeNet-300-100.

\begin{table*}[htb]
\small
\centering
\begin{tabular}{lllllllll}
\hline
\hline
DCNN &Model  & Error (\%) &Channels &Pruned &Params &Pruned   \\
\hline
            &Base-unpruned \cite{liu2017learning}  &6.34 &5504  &-  &20.04M&- \\
            &Pruned \cite{liu2017learning}   &6.20 &1651  &70\%  &2.30M &88.5\% \\
VGG-16      &Base-unpruned (Ours)  &6.34 &4224  &-  &14.98M&- \\
            &Pruned (Ours)   &\bf{6.14} &1689  &60\%  &4.40M&70.6\% \\
            &Pruned (Ours)   &\bf{6.20} &1267  &70\%  &2.88M&80.7\% \\
\hline
            &Base-unpruned \cite{liu2017learning}  &6.11 &9360  &-  &1.02M&- \\
DenseNet-40  &Pruned \cite{liu2017learning}   &5.65 &2808  &70\%  &0.35M&65.2\% \\
            &Pruned (Ours)   &\bf{5.48} &3744  &60\%  &0.45M&55.9\% \\
            &Pruned (Ours)   &\bf{5.57} &2808  &70\%  &0.34M&66.7\% \\
\hline
\hline
\vspace{+1mm}
\end{tabular}
\caption{Pruning performance results comparison on CIFAR-10.}
\label{tab:CIFAR10}
\end{table*}

\begin{table*}[htb]
\small
\centering
\begin{tabular}{lllllllll}
\hline
\hline
DCNN &Model  & Error (\%) &Channels &Pruned &Params &Pruned   \\
\hline
            &Base-unpruned \cite{liu2017learning}  &26.74 &5504  &-  &20.08M&- \\
VGG-16      &Pruned \cite{liu2017learning}   &26.52 &2752  &50\%  &5.00M&75.1\% \\
            &Base-unpruned (Ours)  &26.81 &4224  &-  &15.02M&- \\
            &Pruned (Ours)   &26.55 &2112  &50\%  &6.01M&60.0\% \\
\hline
            &Base-unpruned \cite{liu2017learning}  &25.36 &9360  &-  &1.06M&- \\
DenseNet-40  &Pruned \cite{liu2017learning}   &25.72 &3744  &60\%  &0.46M&54.6\% \\
            &Pruned (Ours)   &\bf{25.66} &3744  &60\%  &0.47M&55.6\% \\
\hline
\hline
\vspace{+1mm}
\end{tabular}
\caption{Pruning performance results comparison on CIFAR-100.}
\label{tab:CIFAR100}
\end{table*}

\subsection{Structured Channel Pruning on the CIFAR and SVHN Datasets}

The CIFAR datasets (CIFAR10 and CIFAR100) provided by  \cite{krizhevsky2009learning} are well established computer vision datasets used for image classification and object recognition. 
Both CIFAR datasets consist of a total of 60K natural color images and are divided into a training dataset with 50K images and a testing dataset with 10K images. 
The CIFAR-10 dataset is drawn from 10 classes with 6000 images per class. 
The CIFAR-100 dataset is drawn from 100 classes with 60 images per class. 
The color images in the CIFAR datasets have resolution $32\times32$.

The SVHN dataset \cite{netzer2011reading} is a real-world image dataset for developing machine learning classification and object recognition algorithms.
Similar to MNIST it consists of cropped digit images, but has as many as 600K training samples and 26K testing images in total. 
Each digit image is $32\times32$ and extracted from natural scenes. 

In this section, we will test our structured channel pruning method \textbf{DSC-2} on two network models: VGG-16 \cite{simonyan2014very} and DenseNet40 \cite{huang2017densely}. 
The VGG-16 \cite{simonyan2014very} is a deep convolutional neural network containing 16 layers which was mainly designed for the ImageNet dataset.
Here we adopt a variation of VGG-16 designed for CIFAR datasets, which was used in \cite{li2016pruning} and has a smaller number of total parameters compared to Liu's \cite{liu2017learning}, to conduct our experiments and compare with other state of art pruning algorithms. 
For DenseNet \cite{huang2017densely} we adopted the DenseNet40 with a total of 40 layers and a growth rate of 12.

We first train all the networks from scratch to obtain similar baseline results compared to \cite{liu2017learning}. 
The total epochs for training was set to 250 epochs for CIFAR, 20 epochs for SVHN, for all networks. The batch size used was 128. 
A Stochastic Gradient Descent (SGD) optimizer with an initial learning rate of 0.1, weight decay of $5\times10^{-4}$ and momentum of $0.9$ was adopted. 
A division of the learning rate by 5 occurs at every $25\%, 50\%, 75\%$ of total training epochs. For these datasets, standard data augmentation techniques like normalization, random flipping, and cropping may be applied.

During the pruning and fine-tuning procedure, the same number of training epochs is adopted in total. 
We use an SGD optimizer with an initial learning rate of $0.005$ and no weight decay or very small weight decay for pruning and fine-tuning purposes.
Similarly, a division of the learning rate by 2 occurs at every $25\%, 50\%, 75\%$ of total training epochs.
For the annealing schedule, a grid search is utilized here to determine the best $p_0$, $N^{c}$ and $\alpha$ for different $p$. 
After the first part of the pruning schedule when we reach the pruning target $p_0$, we conduct the fine-grain pruning for each final pruning rate $p$. 
We output our best results in Table \ref{tab:CIFAR10} for CIFAR 10, Table \ref{tab:CIFAR100} for CIFAR 100 and Table \ref{tab:SVHN} for SVHN.

\begin{table*}[htb]
\small
\centering
\begin{tabular}{lllllllll}
\hline
\hline
DCNN &Model  & Error (\%) &Channels &Pruned &Params &Pruned   \\
\hline
            &Base-unpruned \cite{liu2017learning}  &2.17 &5504  &-  &20.04M&- \\
            &Pruned \cite{liu2017learning}   &\bf{2.06} &2201  &60\%  &3.04M &84.8\% \\
VGG-16      &Base-unpruned (Ours)  &2.18 &4224  &-  &14.98M&- \\
            &Pruned (Ours)   &\bf{2.06} &1689  &60\%  &4.31M&71.2\% \\
\hline
            &Base-unpruned \cite{liu2017learning}  &1.89 &9360  &-  &1.02M &- \\
DenseNet-40  &Pruned \cite{liu2017learning}   &1.81 &3744  &60\%  &0.44M &56.6\% \\
            &Pruned (Ours)   &\bf{1.80} &3744  &60\%  &0.46M&54.9\% \\
\hline
\hline
\vspace{+1mm}
\end{tabular}
\caption{Pruning performance results comparison on SVHN.}
\label{tab:SVHN}
\end{table*}

\begin{figure*}[htb]
\centering
\includegraphics[width=5.9cm]{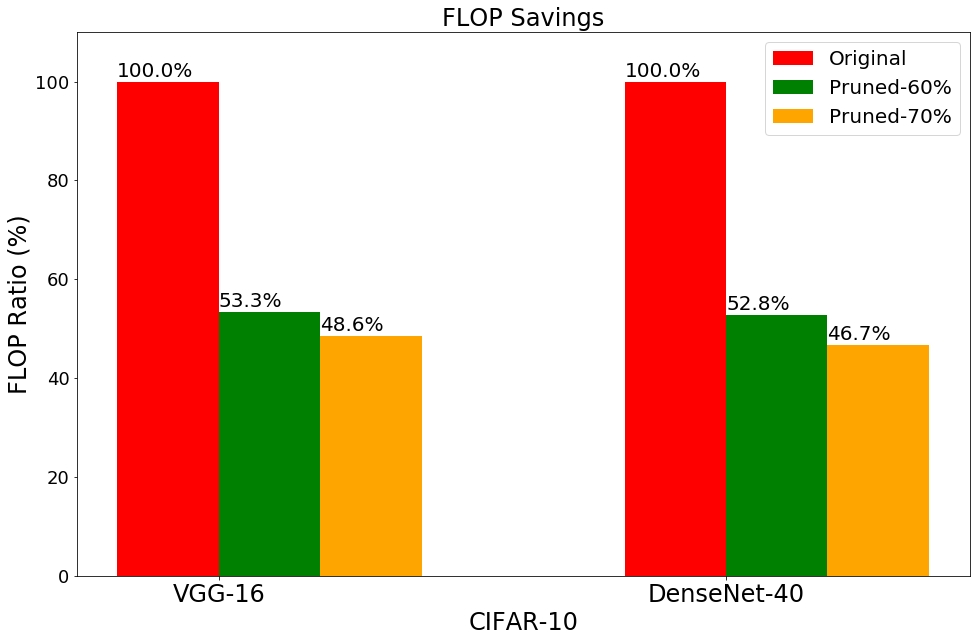}
\includegraphics[width=5.9cm]{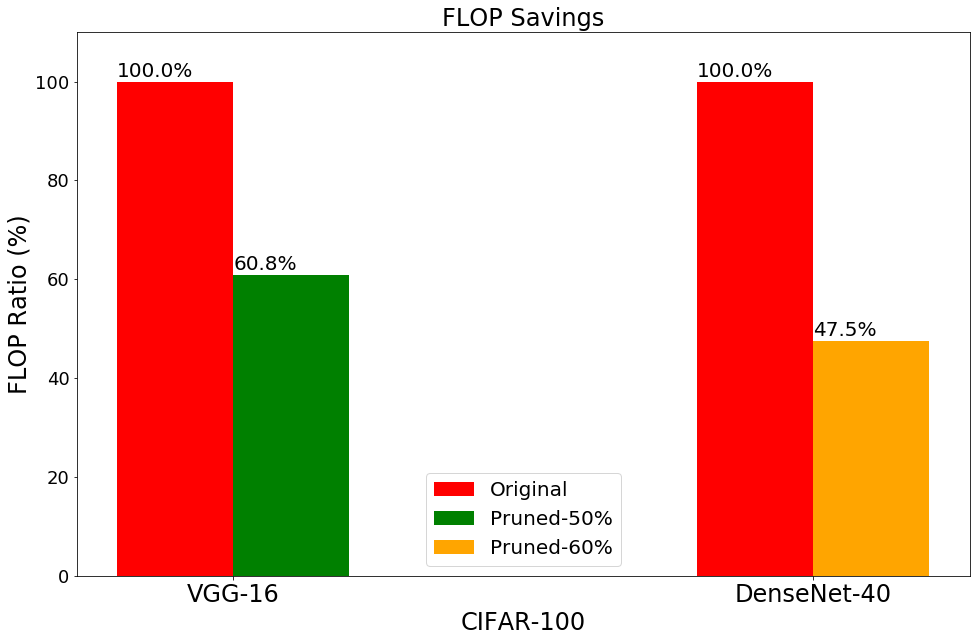}
\includegraphics[width=5.9cm]{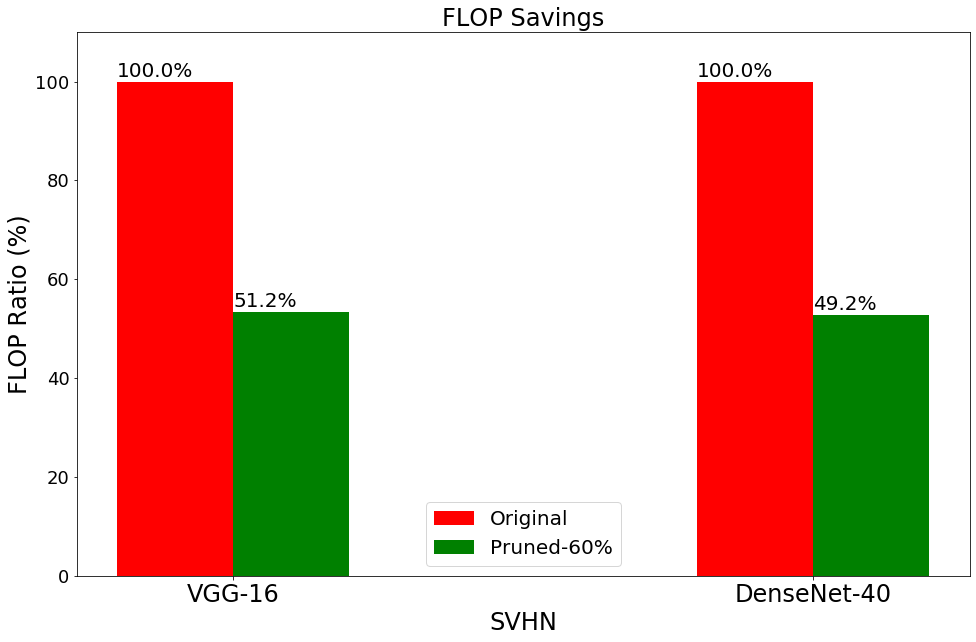}
\caption{FLOPs ratio between the original DCNNs and pruned sub-network for VGG-16 and DenseNet-40 on CIFAR and SVHN dataset.}\label{fig:FLOPs_ratio}
\end{figure*}

The experimental results displayed in Tables \ref{tab:CIFAR10}, \ref{tab:CIFAR100} and \ref{tab:SVHN} demonstrate the effectiveness of our proposed channel pruning algorithm \textbf{DSC-2}. 
It can be observed that our \textbf{DSC-2} method can obtain results competitive with or even better than  \cite{liu2017learning}. 
What's even better, our \textbf{DSC-2} pruning method does not introduce any extra term in the training loss function. 
By using the annealing schedule to gradually remove the "unimportant" channels based on a specified channel importance ranking metric $\mathcal{R}$, we could successfully find a compact sub-network without losing any model performance. 
Our \textbf{DSC-2} is easy to use and can be easily scaled up to any untrained or existing pre-trained model. The results of the FLOPs ratio between the original DCNNs and pruned sub-networks are shown in Figure \ref{fig:FLOPs_ratio}.

\begin{figure*}[htb]
\centering
\begin{tabular}{cc}
\includegraphics[width=7.8cm]{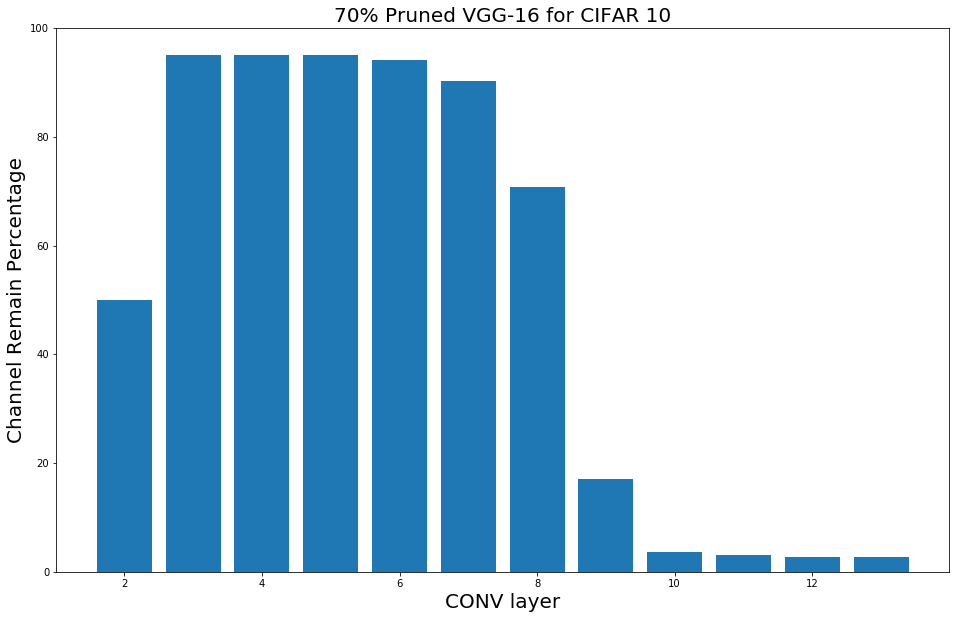}
&\includegraphics[width=7.8cm]{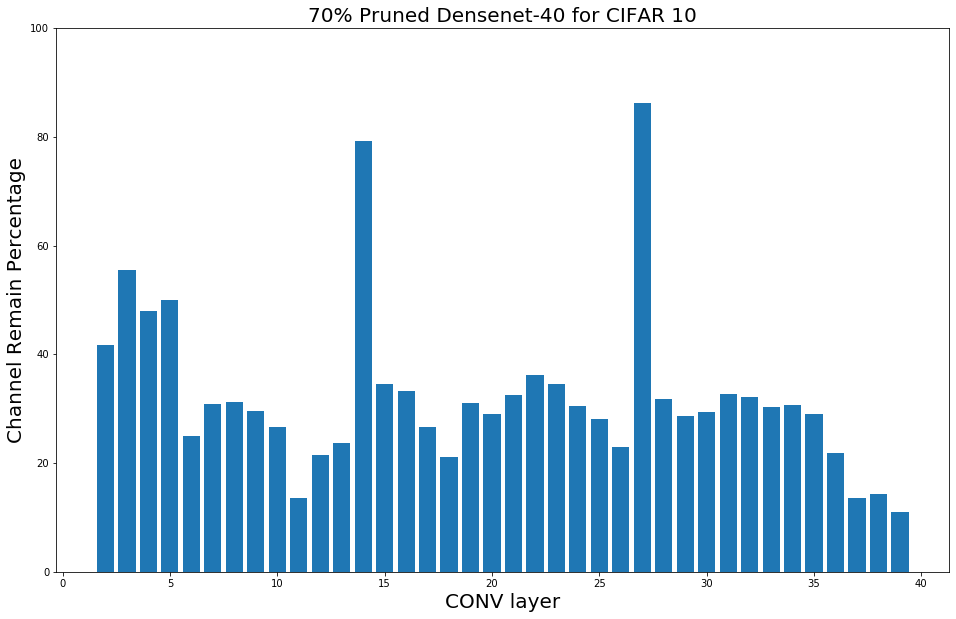}
\\ $70\%$ channel pruned VGG-16
& $70\%$ channel pruned DenseNet-40
\end{tabular}
\caption{The remaining channel distribution for each CONV layer for pruned networks on CIFAR 10. The first CONV layer is not displayed as our channel pruning algorithm \textbf{DSC-2} will not act on this layer, thus contain the full percentage of channels.}\label{fig:channel_distribution}
\end{figure*} 

Figure \ref{fig:channel_distribution} displays two 70\% channel-pruned network models for the CIFAR-10 dataset. 
Due to the significant differences in network architecture between the VGG-16 and DenseNet-40, the resulting distribution of the percentage of remaining channels is quite different. 
For VGG-16, only a very small number of channels are kept in the last five CONV layers. This is reasonable as the last five CONV layers are those layers that initially have 512 input channels. 
Evidently, we do not need so many channels in each of the last five layers. The high pruning percentage may suggest that the VGG-16 network is over-parameterized in a layer-wise way for the CIFAR 10 dataset. 
For DenseNet-40 with a growth rate of 12, the kept channel percentage is relatively evenly distributed in each CONV layer except the two transitional layers. 
This is again very reasonable based on the special architecture of DenseNet. 
With a growth rate of 12, every 12 consecutive layers are correlated with each other, and outputs of those previous CONV layers will be concatenated to be the inputs of the following CONV layer inside the growth rate period. 
Only the transitional layers do not hold that property. 
Overall, our channel-level pruning algorithm \textbf{DSC-2} can automatically detect the reasonable sub-network without performance loss for VGG-16 and DenseNet-40 on the CIFAR and SVHN datasets.

\begin{figure*}[htb]
\centering
\begin{tabular}{cc}
\includegraphics[width=7.8cm]{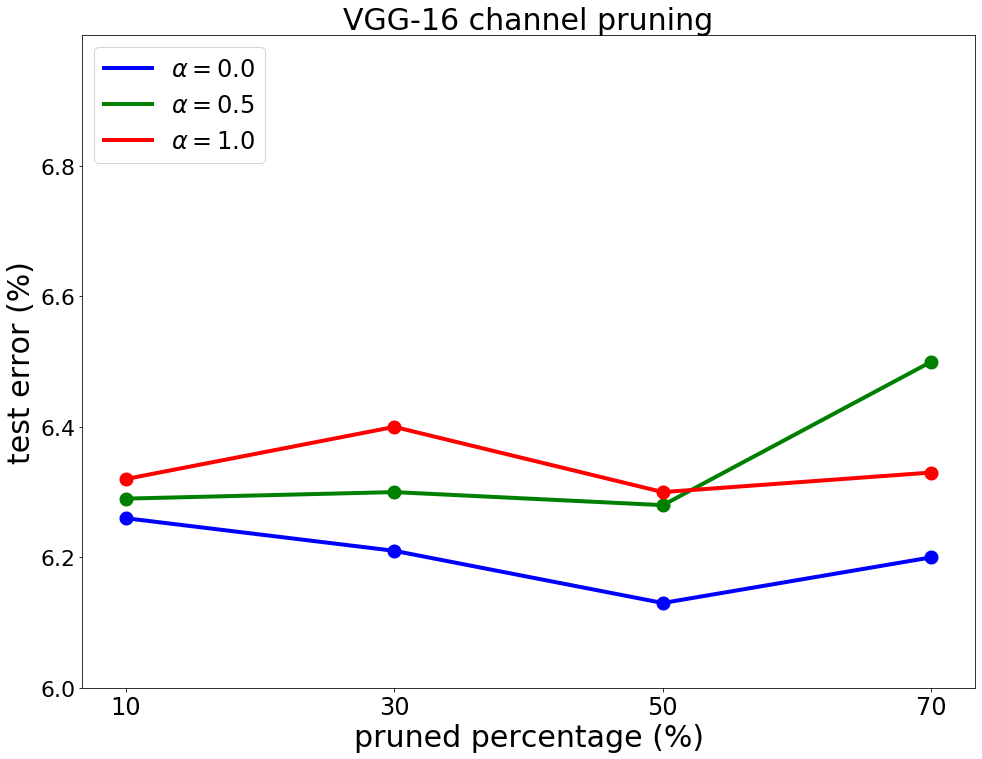}
&\includegraphics[width=7.8cm]{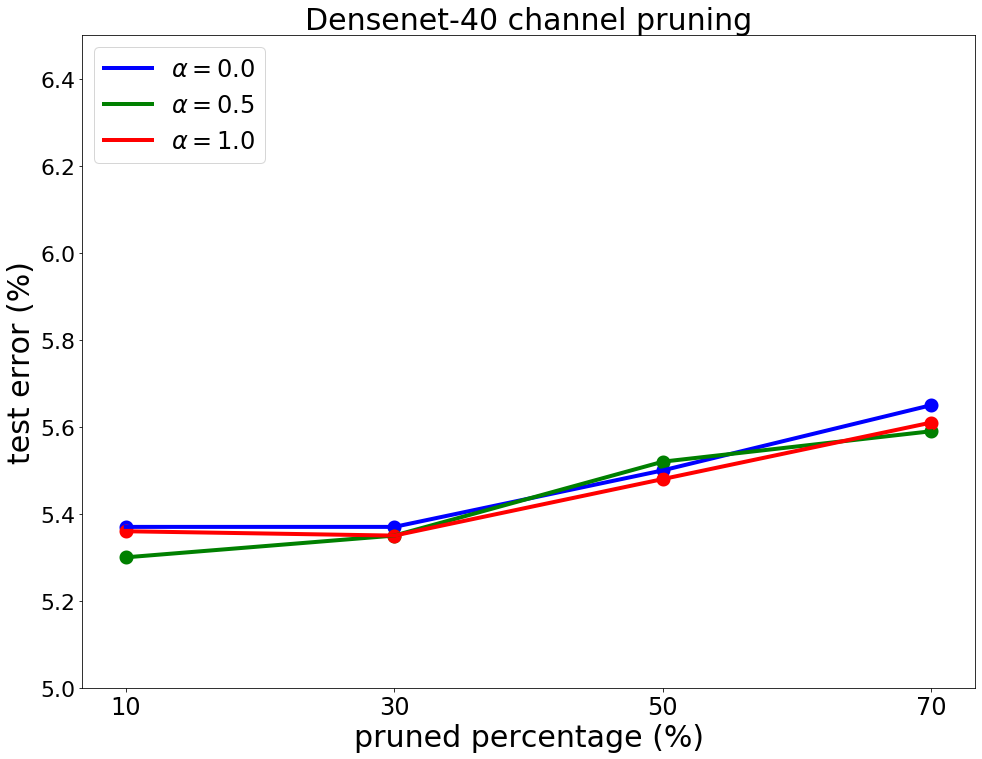}
\\ VGG-16
& DenseNet-40
\end{tabular}
\caption{The test error for various channel pruned percentage using one single global pruning rate for CIFAR-10.}\label{fig:distribution_percentage}
\end{figure*} 

Figure \ref{fig:distribution_percentage} displays the best results we obtained for just using one single global pruning rate $p \in \{0.1, 0.3, 0.5, 0.7 \}$ to perform the channel pruning in the whole channel space. 
We can observe that even using a single target pruning rate parameter, we can still obtain very good sub-networks that generalize to CIFAR-10 test data well. 
A large portion of our best results displayed in Tables \ref{tab:CIFAR10}, \ref{tab:CIFAR100} and \ref{tab:SVHN} are obtained just using a single global pruning rate to guide the network pruning procedure. 
This observation makes our annealing pruning algorithm very easy to use, without worrying about channel subspace partitions. 
We also tested three different values for the parameter $\alpha$ of Eq. \eqref{eq:R_structured}. For DenseNet-40, all of the choices of $\alpha$ can yield satisfactory performance for the pruned sub-network. For VGG-16, when $\alpha$ is set to 0, that is when the magnitude of $\gamma$ is used as the ranking metric for channel pruning, gives the best results. This implies that different $\alpha$ values may be suitable for different network architectures.

\section{Conclusion}

This paper presented a neural network pruning framework that is suitable for both structured and non-structured pruning. 
The method directly imposes a $L_0$ sparsity constraint on the network parameters, which is gradually tightened to the desired sparsity level. 
This direct control allows us to obtain the precise sparsity level desired, as opposed to other methods that obtain the sparsity level indirectly through either a quality factor times the variance or the use of penalty parameters.
Experiments on extensive synthetic and real vision data, including the MNIST, CIFAR, and SVHN provide empirical evidence that the proposed network pruning scheme obtains a performance comparable to or better than other state of art pruning methods.

\bibliographystyle{IEEEtran}
\bibliography{IEEEexample}

\end{document}